%% file: root.tex
\documentclass[letterpaper, twoside, 10 pt, journal]{IEEEtran}
\usepackage{orcidlink}
\usepackage{amsmath} % assumes amsmath package installed
\usepackage{amssymb}  % assumes amsmath package installed
\usepackage{multirow} % Tabelle con multiriga?
\usepackage{array}
\usepackage{textcomp}
\usepackage{stfloats}
\usepackage{siunitx}
\usepackage{gensymb}
\usepackage{textcomp}
\usepackage{cite}
\usepackage{tikz}
\usepackage{rotating} 
\usepackage{tablefootnote}

\usepackage{tabularray}

\usepackage{tikz}

\newcommand\copyrighttext{%
  \footnotesize \textcopyright \the\year{} IEEE. Personal use of this material is permitted. Permission from IEEE must be obtained for all other uses, including reprinting/republishing this material for advertising or promotional purposes, collecting new collected works for resale or redistribution to servers or lists, or reuse of any copyrighted component of this work in other works.}

\title{\LARGE \bf
MineInsight: A Multi-sensor Dataset for Humanitarian Demining Robotics in Off-Road Environments 
}

\author{Mario Malizia~\orcidlink{0009-0008-6618-5059}, Charles Hamesse~\orcidlink{0000-0002-2321-0620}, Ken Hasselmann~\orcidlink{0000-0002-8196-9889}, \\Geert De Cubber~\orcidlink{0000-0001-7772-0258}, Nikolaos Tsiogkas~\orcidlink{0000-0003-2842-7316}, Eric Demeester~\orcidlink{0000-0001-6866-3802}, Rob Haelterman~\orcidlink{0000-0002-1610-2218}

\thanks{Manuscript received: March 20, 2025; Revised July 31, 2025; Accepted November 8, 2025.}%Use only for final RAL version
\thanks{This paper was recommended for publication by Editor Hyungpil Moon upon evaluation of the Associate Editor and Reviewers’ comments.} 
\thanks{This work was supported by the Belgian Defense under Grant DAP 23/08. (Corresponding author: Mario Malizia) - Email: \textit{mario.malizia@mil.be}.}
\thanks{M. Malizia is with the Department of 
Mechanics, Royal Military Academy, Av. de la Renaissance 30, 1000 Brussels, Belgium, and also with the ACRO Research Group, Department of Mechanical Engineering, Wetenschapspark 27, 3590 Diepenbeek, 
Belgium, and Flanders Make @ KU Leuven, B-3001 Heverlee, Belgium.}%
\thanks{C. Hamesse and R. Haelterman are with the Department of Mathematics, Royal Military Academy, Av. de la Renaissance 30, 1000 Brussels, Belgium.}%
\thanks{K. Hasselmann and G. De Cubber are with the Department of 
Mechanics, Royal Military Academy, Av. de la Renaissance 30, 1000 Brussels, Belgium.}% 
\thanks{N. Tsiogkas is with the ACRO Research Group, Department of Computer Science, KU Leuven, Wetenschapspark 27, 3590 Diepenbeek, 
Belgium, and also with Flanders Make @ KU Leuven, B-3001 Heverlee, Belgium.}% 
\thanks{E. Demeester is with the ACRO Research Group, Department of Mechanical Engineering, KU Leuven, Wetenschapspark 27, 3590 Diepenbeek, Belgium, and also with Flanders Make @ KU Leuven, B-3001 Heverlee, Belgium.}
\thanks{Digital Object Identifier (DOI): see top of this page.}}%

\begin{document}

\IEEEoverridecommandlockouts
\maketitle
%\thispagestyle{empty}
%\pagestyle{empty}

%\submittednotice

%%%%%%%%%%%%%%%%%%%%%%%%%%%%%%%%%%%%%%%%%%%%%%%%%%%%%%%%%%%%%%%%%%%%%%%%%%%%%%%%
\markboth{IEEE Robotics and Automation Letters. Preprint Version. Accepted December, 2025}
{Malizia \MakeLowercase{\textit{et al.}}: MineInsight: A Multi-sensor Dataset for Humanitarian Demining Robotics in Off-Road Environments}

%%%%%% ABSTRACT BEGINS %%%%%%
\input{0_abstract}
%%%%%% ABSTRACT ENDS %%%%%%
%%%%%%%% INDEX TERMS IN THE ABSTRACT %%%%%%%
%%%%%%%%%%%%%%%%%%%%%%%%%%%%%%%%%%%%%%%%%%%%%%%%%%%%%%%%%%%%%%%%%%%%%%%%%%%%%%%%
\input{1_introduction}
\input{2_related_work}

\input{3_sensor_setup_and_calibration}
\input{4_dataset_description}

\input{5_baseline_evaluation}
\input{6_limitations}
\input{7_conclusion}
\section*{ACKNOWLEDGMENT}
The authors thank Alessandra Miuccio and Timothée Fréville for their support in the hardware and software design. They also thank Sanne Van Hees and Jorick Van Kwikenborne for their assistance in organizing the measurement campaign. Finally, they thank the Belgian Meteo Wing for providing the climatology study during the days of the test campaign. 
%%%%%%%%%%%%%%%%%%%%%%%%%%%%%%%%%%%%%%%%%%%%%%%%%%%%%%%%%%%%%%%%%%%%%%%%%%%%%%%%
\bibliographystyle{IEEEtran}
\bibliography{paper_bib}  

\end{document}

%% file: 0_abstract.tex
\begin{abstract}
The use of robotics in humanitarian demining increasingly involves computer vision techniques to improve landmine detection capabilities.
However, in the absence of diverse and realistic datasets, the reliable validation of algorithms remains a challenge for the research community.
In this paper, we introduce MineInsight, a publicly available multi-sensor, multi-spectral dataset designed for off-road landmine detection. The dataset features 35 different targets (15 landmines and 20 commonly found objects) distributed along three distinct tracks, providing a diverse and realistic testing environment.
MineInsight is, to the best of our knowledge, the first dataset to integrate dual-view sensor scans from both an Unmanned Ground Vehicle and its robotic arm, offering multiple viewpoints to mitigate occlusions and improve spatial awareness.
It features two LiDARs, as well as images captured at diverse spectral ranges, including visible (RGB, monochrome), visible short-wave infrared (VIS-SWIR), and long-wave infrared (LWIR).
Additionally, the dataset provides bounding boxes generated by an automated pipeline and refined with human supervision.
We recorded approximately one hour of data in both daylight and nighttime conditions, resulting in around 38,000 RGB frames, 53,000 VIS-SWIR frames, and 108,000 LWIR frames.
MineInsight serves as a benchmark for developing and evaluating landmine detection algorithms.
Our dataset is available at~\url{https://github.com/mariomlz99/MineInsight}.
\end{abstract}

\begin{IEEEkeywords}
Data sets for robotic vision, field robotics, computer vision for
automation.
\end{IEEEkeywords}

%% file: 1_introduction.tex
\section{Introduction}
\IEEEPARstart{L}{andmines} remain one of the most dangerous legacies of conflicts, posing long-term threats to civilians and obstructing post-war recovery efforts. In many affected regions, the presence of antipersonnel (AP) and antitank (AT) landmines severely limits agricultural activities, infrastructure development, and safe mobility. Traditional humanitarian demining methods, relying on human experience and metal detectors, are slow, hazardous, and costly.
Consequently, robotic systems have been increasingly deployed~\cite{nevliudov_robots} to accelerate landmine detection and improve operational safety. 
While many robotic systems rely solely on camera feedback for operators, researchers have increasingly explored diverse detection techniques~\cite{popov_uav,baur_dl_uav,colreavy_cnn,winfred_thermal_cnn,milan_uav_uxo_det,dnn_landmine} exploiting images captured across multiple spectral ranges.
%%%  
Existing landmine detection datasets~(Table \ref{tab:datasets_comparisons}) often suffer from significant limitations, such as a narrow range of explosive ordnance types, limited sensor diversity, or inadequate coverage of challenging environmental conditions. These constraints on existing datasets hinder the development and validation of novel detection algorithms. 
Moreover, the lack of a standardized storage format for data releases makes it difficult to ensure consistency and comparability across studies, highlighting the need for datasets that follow standardized practices established in off-road mobile robotics datasets~\cite{rugd_dataset,rellis3d_dataset,citrus_dataset}.
%%%   
Most datasets rely on one sensor, neglecting sensor fusion applications incorporating other spectral ranges such as visible short-wave infrared (VIS-SWIR) or long-wave infrared (LWIR).
Additionally, these datasets often omit LiDAR data, which delivers essential depth information for navigating uneven terrain, detecting obstacles, and improving landmine localization.
They do not adequately address occlusion, a major challenge when mines are covered with vegetation, debris, or soil~\cite{baur_occlusion}. To the best of our knowledge, the majority of existing datasets in the literature are recorded using Unmanned Aerial Vehicles (UAVs). However, in many scenarios, deploying Unmanned Ground Vehicles (UGVs) is more practical or necessary, highlighting a gap that we aim to fill.
%%%
To address these gaps, we introduce \textbf{MineInsight}, a multi-sensor, multi-spectral dataset designed to advance robotic landmine detection.
MineInsight integrates dual sensor perspectives, capturing both 2D and 3D information from a UGV and its robotic arm, enabling a wider spatial understanding of landmines in complex terrains.
Our dataset features two LiDARs and a diverse set of multi-spectral sensors spanning monochrome, RGB, VIS-SWIR, and LWIR spectra.

\noindent{The contributions of this paper are as follows:}
\begin{itemize}
    \item We introduce MineInsight, a dataset featuring a diverse array of inert AP and AT landmines, along with common items that might be found in natural environments, to better mimic realistic demining scenarios and challenge detection systems against false positives.
    \item Our dataset is, to the best of our knowledge, the first dataset that integrates dual sensor scans captured from both the UGV and its robotic arm. The robotic arm’s dynamic motion provides varying sensor viewpoints that effectively mitigate obstructions and enhance the detection of hidden landmines.
    \item We provide 2D bounding boxes refined with human supervision, serving as a benchmark for validating detection algorithms and supporting further research.
\end{itemize}

%% file: 2_related_work.tex
\section{Related Work}
\label{sec_related_work}
%%%%%%%%%%%%%%%%%%%%%%%%%
Landmine detection datasets, as illustrated in Table~\ref{tab:datasets_comparisons}, are relatively rare in the existing literature.
%%%%%%%%%%%%%%%%%%%%%%%%%
%%% BEGIN TABLE ON EXISTING DATASETS %%%
\begin{table*}[t]
\caption{Comparison of different landmine detection datasets. Missing entries indicate unavailable data.
\\\textit{Note: “AVision” is the short form for AlliedVision.}}
\label{tab:datasets_comparisons}
\begin{tabular}{lcccccc}
\hline
\textbf{Dataset}                        & \textbf{\begin{tabular}[c]{@{}c@{}}Total\\ Landmines\end{tabular}} & \textbf{Platform}                                           & \textbf{Environment}                                                        & \textbf{Camera}                                                                                                                                           & \textbf{LiDAR}                                                      & \textbf{Labels}                                                                                             \\ \hline
De Smet et al.~\cite{scatterable_1_7}   & 1                                                                  & UAV                                                         & Grass, rubble                                                               & \begin{tabular}[c]{@{}c@{}}Parrot Sequoia \\ Multispectral\end{tabular}                                                                                   & --                                                                  & --                                                                                                          \\
Steinberg et al.~\cite{scatterable_8_9} & 1                                                                  & UAV                                                         & Grass, rubble                                                               & \begin{tabular}[c]{@{}c@{}}Parrot Sequoia \\ Multispectral\end{tabular}                                                                                   & --                                                                  & \begin{tabular}[c]{@{}c@{}}2D \\ bounding boxes\end{tabular}                                                  \\
Baur et al.~\cite{scatterable_10_22}    & 1                                                                  & UAV                                                         & Sand                                                                        & \begin{tabular}[c]{@{}c@{}}Parrot Sequoia \\ Multispectral\end{tabular}                                                                                   & --                                                                  & --                                                                                                          \\
Leloglu et al.~\cite{leloglu_dataset}   & 1                                                                  & \begin{tabular}[c]{@{}c@{}}Fixed \\ Tripod\end{tabular}     & Clay                                                                        & \begin{tabular}[c]{@{}c@{}}ATOM 1024 IR,\\ FLIR T 650 SC\end{tabular}                                                                                     & --                                                                  & --                                                                                                          \\
Tamayo et al.~\cite{tamayo_dataset}     & 1                                                                  & UAV                                                         & \begin{tabular}[c]{@{}c@{}}Low vegetation,\\ clay\end{tabular}              & Zenmuse XT IR                                                                                                                                             & --                                                                  & --                                                                                                          \\
Vivoli et al.~\cite{vivoli_yolo}        & 2                                                                  & \begin{tabular}[c]{@{}c@{}}Handheld \\ Support\end{tabular} & Grass, gravel                                                               & iPhone 13                                                                                                                                                 & --                                                                  & \begin{tabular}[c]{@{}c@{}}2D \\ bounding boxes\end{tabular}                                                  \\
\textbf{MineInsight (Ours)}             & 15                                                                 & UGV                                                         & \begin{tabular}[c]{@{}c@{}}Low to high \\ vegetation,\\ leaves\end{tabular} & \begin{tabular}[c]{@{}c@{}}AVision Alvium 1800 U-240C,\\ AVision Alvium 1800 U-130,\\ Sevensense Core Research,\\ Teledyne FLIR Boson 640\end{tabular} & \begin{tabular}[c]{@{}c@{}}Livox AVIA,\\ Livox Mid-360\end{tabular} & \begin{tabular}[c]{@{}c@{}}2D \\ bounding boxes
\end{tabular} \\ \hline
\end{tabular}
\end{table*}
%%% END TABLE ON EXISTING DATASETS %%%
%%%%%%%%%%%%%%%%%%%%%%%%%
This section offers an overview of the available ones, detailing their content and the conditions under which they were collected.

Binghamton University has contributed significantly to the field of landmine detection by releasing multiple datasets as part of the Scatterable Landmine Detection Project.
These datasets include data from various UAV flights flown over different terrains.
The project's primary focus is the detection of PFM-1 landmines, together with their carrying case.
We propose a categorization of the dataset releases into three groups (1–7, 8–9, and 10–24), facilitating a direct comparison between UAV platforms, sensors, and data collection methods across studies.
The first group (Datasets 1-7), released by De Smet et al.~\cite{scatterable_1_7}, includes data collected using a DJI Phantom 4 Pro UAV and a DJI Matrice 600 Pro UAV, both equipped with a Parrot Sequoia Multispectral Camera. Flights were conducted over grass and rubble environments, resulting in seven high-resolution orthophotos.
Steinberg et al. released the second group (Datasets 8-9)~\cite{scatterable_8_9}, which includes data collected from three flights over grass and four flights over rubble using a DJI Matrice 600 Pro UAV equipped with a Parrot Sequoia Multispectral Camera. Baur et al. later used the same data to explore the potential of deep learning-based landmine detection, with their experimental setup described in~\cite{baur_dl_uav}, where they address the challenges of detecting PFM-1 landmines in UAV-based imagery.
Baur et al.~\cite{scatterable_10_22} released the third group (Datasets 10-24), which comprises photogrammetric data collected using a DJI Phantom 4 UAV equipped with a Parrot Sequoia Multispectral Camera and processed with Pix4DMapper. 
To introduce variability, the PFM-1 landmines were rotated, partially buried, or partially hidden in each orthophoto.

Kaya and Leloglu~\cite{kaya_thermal} conducted research on the detection of various landmine types, including M15, M16, M2, M48 metal, DM-11, and M14. These landmines, filled with wax or metal to simulate real thermal properties, were placed at varying depths (surface, semi-buried, and buried) to replicate real-world conditions.
Experiments were conducted in two test areas with different soil types (clay and quasi-humid clay loam), cleared of vegetation to reduce thermal noise.
As part of their study, the authors released a dataset~\cite{leloglu_dataset} comprising 96 thermal images captured at 15-minute intervals over 24 hours. Data were collected with the ATOM 1024 infrared (IR) and FLIR T 650 SC cameras, which have spectral ranges of 8000–14000 ~\si{nm} and 7500–13000 ~\si{nm}, respectively. The cameras were positioned at fixed altitudes on a roof, pointing toward the test areas. However, the resulting dataset only includes DM-11 landmines.

Tamayo et al.~\cite{tamayo_dataset} released a dataset for landmine detection containing thermographic and visible spectrum images of buried landmines at varying depths. The dataset includes 2,700 thermographic images, specifically designed to detect ``legbreaker" antipersonnel mines. These landmines were custom replicas, made from a PVC cylinder, a \SI{5}{ml} syringe as a detonator, \SI{20}{g} of nails, and \SI{400}{g} of anthracite charcoal, chosen for its thermal properties similar to TNT.
The experiments were conducted on a terrain with a low percentage of vegetation and clay soil. The mines were buried at various depths, and distributed across distinct zones to facilitate segmentation in image analysis.
The dataset was collected using a Zenmuse XT IR camera with a spectral range of 7500–13000~\si{nm}, mounted on a DJI Matrice 100 UAV. Environmental parameters such as temperature, irradiance, wind speed, and relative humidity were recorded alongside the images to provide context for thermal variations in the dataset.

Vivoli et al.~\cite{vivoli_yolo} released SULAND, a dataset for surface landmine detection. It consists of 33,541 RGB images captured using an iPhone 13 camera, with 10,169 images annotated for object detection tasks.
The sequences were collected under varying weather conditions, including cloudy, sunny, and shadowed settings, across grass and gravel environments. The targets selected for this study were the PFM-1 and PMA-2 landmines.
Data were collected in diverse environmental settings, considering factors such as terrain type, weather conditions, and obstacles like bushes, rocks, and branches that could affect visibility.

With the increasing availability of services such as Roboflow~\cite{roboflow_web}, a wide variety of landmine detection datasets have become publicly accessible.
These datasets include both real and simulated images captured using RGB or thermal cameras. However, they often lack essential details about the data collection process, such as camera specifications, environmental conditions, and experimental setup. This omission makes it challenging to replicate experiments or validate algorithms effectively.
%%%%%%%%%%%%%%%%%%%%%%%%%

Existing landmine detection datasets lack a standardized framework for data collection and release. We follow established practices from robotic datasets, adopting the Robot Operating System (ROS) 2 framework to maintain interoperability and reproducibility within the research community.
No publicly available landmine detection dataset, to our knowledge, features a UGV-mounted robotic arm for data collection and features 3D sensors, such as LiDAR. Our dataset introduces these elements while also incorporating 15 landmine types, balanced between AP and AT. 
With the inclusion of items commonly found in vegetation and by capturing data in both daylight and nighttime conditions, we enable the analysis of the robustness and sensitivity of detection algorithms.

%% file: 3_sensor_setup_and_calibration.tex
\section{Hardware setup and calibration}
\label{sec_hw_setup_and_calib}
In this section, we describe the hardware configuration of our UGV, including the sensor suite installed on both the mobile base and the robotic arm.
We then introduce the onboard computing architecture responsible for data processing and storage, outline the synchronization approach, and detail the calibration procedures for camera intrinsics and camera–LiDAR extrinsics.
%%%%%%%%%%%%%%%%%%%%%%%%%
\begin{table*}[t]
\caption{Specifications of the sensors used for our data collection. 
% This table summarizes each sensor's type, mounting location on the UGV (platform or robotic arm), resolution, horizontal and vertical field of view, spectral range, and operational frequency. The dataset comprises multi-spectral cameras covering VIS, SWIR, and LWIR wavelengths, alongside LiDAR sensors. 
\\\textit{Notes: “AVision” is the short form for AlliedVision; “FoV” refers to the camera's field of view; “Ch” denotes the number of channels.}} 
\label{tab:sensor_specs}
\centering
\renewcommand{\arraystretch}{1.3}
\begin{tabular}{lcccccccc}
\hline
\textbf{Sensor}                                       & \textbf{Type}   & \textbf{Location} & \textbf{Resolution} & \multicolumn{2}{c}{\textbf{FoV [°]}} & \multicolumn{1}{l}{\textbf{Ch}} & \textbf{\begin{tabular}[c]{@{}c@{}}Spectral \\ Range [nm]\end{tabular}} & \textbf{Rate [Hz]} \\
                                                      &                 &                   &                     & \textbf{H}        & \textbf{V}       & \multicolumn{1}{l}{}            &                                                                         &                    \\ \hline
Livox Mid-360                                         & LiDAR           & Mobile Base       & -                   & 360.0             & 59.0             & -                               & 905\textsuperscript{$\star$}                                            & 10                 \\
Sevensense Core Research\textsuperscript{\textdagger} & Monochrome Camera     & Mobile Base       & 720 $\times$ 540    & 126.0             & 92.4             & 1                               & 380 - 750                                                               & 15                 \\
AVision Alvium 1800 U-240C                            & RGB Camera      & Robotic Arm       & 1916 $\times$ 1216  & 103.6             & 76.7             & 3                               & 300 - 1100                                                              & 15                 \\
AVision Alvium 1800 U-130                             & VIS-SWIR Camera & Robotic Arm       & 1296 $\times$ 1032  & 22.7              & 18.2             & 1                               & 400 - 1700                                                              & 15                 \\
Livox AVIA                                            & LiDAR           & Robotic Arm       & -                   & 70.4              & 77.2             & -                               & 905\textsuperscript{$\star$}                                            & 10                 \\
Teledyne FLIR Boson 640                               & LWIR Camera     & Robotic Arm       & 640 $\times$ 512    & 95.0              & 76.0             & 1                               & 8000 - 13500                                                            & 30                 \\ \hline
\end{tabular}
\\ % THIS SPACE STAYS HERE IF SOMEONE TOUCHES IT I KILL HIM
\vspace{2mm}
\footnotesize{$\star$ The 905 \si{nm} value indicates the central wavelength of the LiDAR’s laser emitter, not a spectral range.\\
\textdagger~For simplicity, this sensor is reported in one line, yet it includes five different cameras.}
\end{table*}
%%%%%%%%%%%%%%%%%%%%%%%%%
\subsection{Hardware setup}
\label{subsec_hw_setup}
\begin{figure}[!t]
    \centering
    \includegraphics[width=1\linewidth]{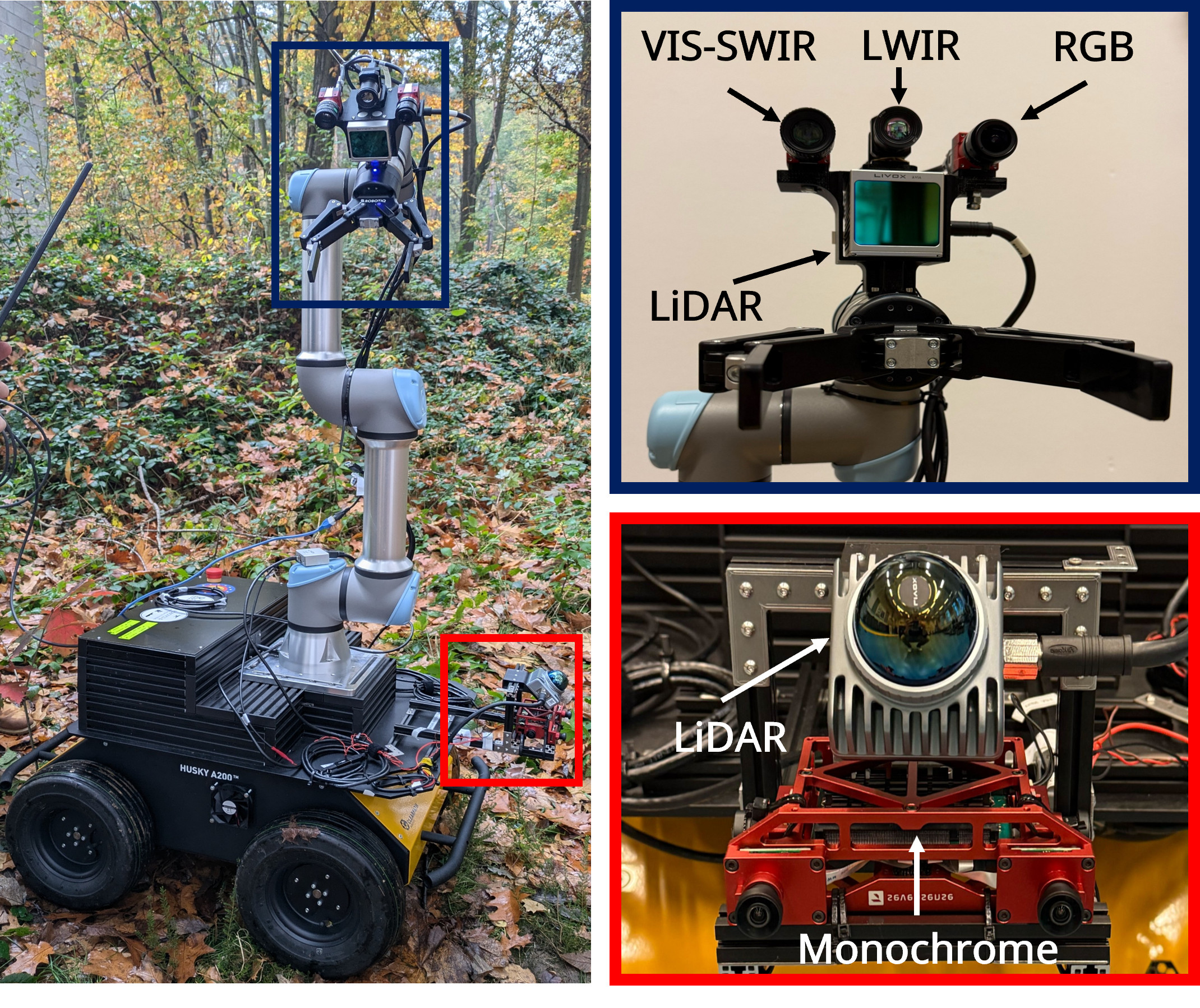}
    \caption{Illustration of the different sensors mounted on the UGV. \textit{Left -} The UGV is ready for data collection. \textit{Top right -} Sensors installed on the robotic arm. \textit{Bottom right:} Sensors mounted directly on the mobile base.}
    \label{fig:sensor_setup}
    \end{figure}

%%%%%%%%%%%%%%%%%%%%%%%%%
\begin{figure*}[!t]
    \centering
    \includegraphics[width=0.75\linewidth]{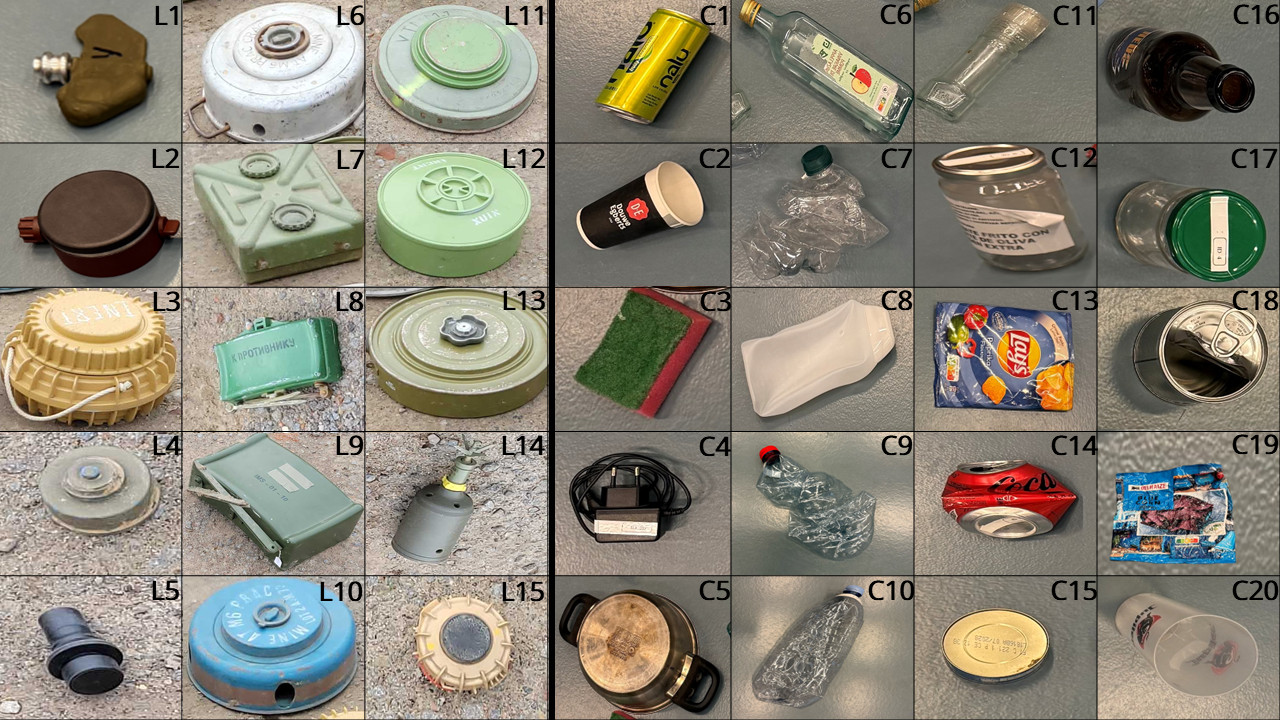}
    \caption{Overview of all the objects included in the dataset. The objects on the left side of the black line are \textbf{landmines (L)}, while the objects on the right side of the black line are \textbf{common items (C)}.
    For what concerns the landmines (L) - \textbf{L1}:~PFM-1 ``Butterfly", \textbf{L2}:~PMN, \textbf{L3}:~TC-3.6, \textbf{L4}:~M35, \textbf{L5}:~C-3~``Elsie", \textbf{L6}:~M6~(grey), \textbf{L7}:~TMA-2, \textbf{L8}:~MON-50, \textbf{L9}:~MON-90, \textbf{L10}:~M6~(blue), \textbf{L11}:~TMM-1, \textbf{L12}:~Type~72~(P), \textbf{L13}:~TM-46, \textbf{L14}:~PROM-1, \textbf{L15}:~VS-50.
    For what concerns the common items (C) - \textbf{C1}:~Soda metal can, \textbf{C2}:~Disposable paper cup, \textbf{C3}:~Sponge, \textbf{C4}:~Plastic charger, \textbf{C5}:~Metal pot, \textbf{C6}:~Glass vinegar bottle, \textbf{C7}:~Plastic water bottle (significantly crumpled), \textbf{C8}:~Plastic shampoo bottle, \textbf{C9}:~Plastic water bottle (partially crumpled), \textbf{C10}:~Plastic water bottle (slightly crumpled), \textbf{C11}:~Glass pepper dispenser, \textbf{C12}:~Glass jar (grey cover), \textbf{C13}:~Plastic chips bag (slightly crumpled), \textbf{C14}:~Metal coke can, \textbf{C15}:~Metal tuna can, \textbf{C16}:~Glass beer bottle, \textbf{C17}:~Glass jar (green cover), \textbf{C18}:~Metal corn tin, \textbf{C19}:~Plastic chips bag (significantly crumpled), \textbf{C20}:~Plastic cup.}
    \label{fig:all_targets}
\end{figure*}
%%%%%%%%%%%%%%%%%%%%%%%%%
The UGV used for data collection, along with its sensor setup, is illustrated in Figure~\ref{fig:sensor_setup}.
We categorize the sensors based on their location into two groups: the mobile base sensor suite and the robotic arm sensor suite, described in Table~\ref{tab:sensor_specs}.
In addition to the visible sensor suite, we mounted a Microstrain 3DM-GV7-AR IMU inside the UGV.
The system relies on two onboard computers: a MYBOTSHOP Industrial PC, which we designate as our embedded computer, and an NVIDIA Jetson Orin (which we refer to as Orin).
The embedded computer manages the control of the mobile base and robotic arm, as well as the processing of data streams from the base-mounted sensors. Meanwhile, the Orin is dedicated to data storage and processing for the sensors mounted on the robotic arm.
We use a gigabit Ethernet switch to link the two computers and the robotic arm, and we ensure synchronization through hardware Precision Time Protocol (PTP).
Both computers run Ubuntu Linux 22.04 LTS and ROS 2 Humble to ensure a common framework for data collection.
%%%%%%%%%%%%%%%%%%%%%%%%%
\subsection{Synchronization}
\label{subsec_sync}
The Orin, the robotic arm, the two LiDARs, and the monochrome cameras are synchronized with the master clock of the embedded computer using hardware PTP.
The RGB, VIS-SWIR, and LWIR cameras, all equipped with global shutters, are connected to the Orin via USB.
Since the Jetson Orin is PTP hardware-synchronized with the embedded computer, the aforementioned USB-connected sensors remain software-synchronized relative to it.
%%%%%%%%%%%%%%%%%%%%%%%%%
\subsection{Camera calibration}
\label{subsec_cam_calib}
As listed in Table~\ref{tab:sensor_specs}, our setup comprises eight cameras, each requiring a distinct calibration approach based on its optical and operational properties.
We calibrate the RGB and VIS-SWIR cameras using the Kalibr toolbox~\cite{kalibr_1}~\cite{kalibr_2}.
For the LWIR camera, calibration is performed using the MRT Calibration Toolbox~\cite{Schramm2021} and a custom-designed checkerboard pattern with metallic squares, inspired by the approach described in~\cite{thermal_calib}.
For the Sevensense Core Research module, we use the manufacturer's parameters.
%%%%%%%%%%%%%%%%%%%%%%%%%
\subsection{Camera and LiDAR calibration}
\label{subsec_cam_lid_calib}
Accurate extrinsic calibration between sensors ensures precise data fusion from multiple streams.
We organize the calibration into two groups based on sensor placement.
The first group includes the Livox Mid-360 LiDAR and each camera in the Sevensense Core Research module. The second group pairs the Livox AVIA LiDAR calibrated with the RGB, VIS-SWIR, or LWIR camera, respectively. We use the method described in~\cite{extrinsic_calib} to perform targetless calibration, aligning each camera frame with the corresponding LiDAR scan.
%%%%%%%%%%%%%%%%%%%%%%%%%

%% file: 4_dataset_description.tex
\section{Dataset}
%%%%%%%%%%%%%%%%%%%%%%%%%
\begin{figure*}[!t]
    \centering
    \begin{tikzpicture}
        % -- Main image with two rows (top row + bottom row)
        \node[anchor=south west, inner sep=0] (image) at (0,0) {%
            \includegraphics[width=0.75\linewidth]{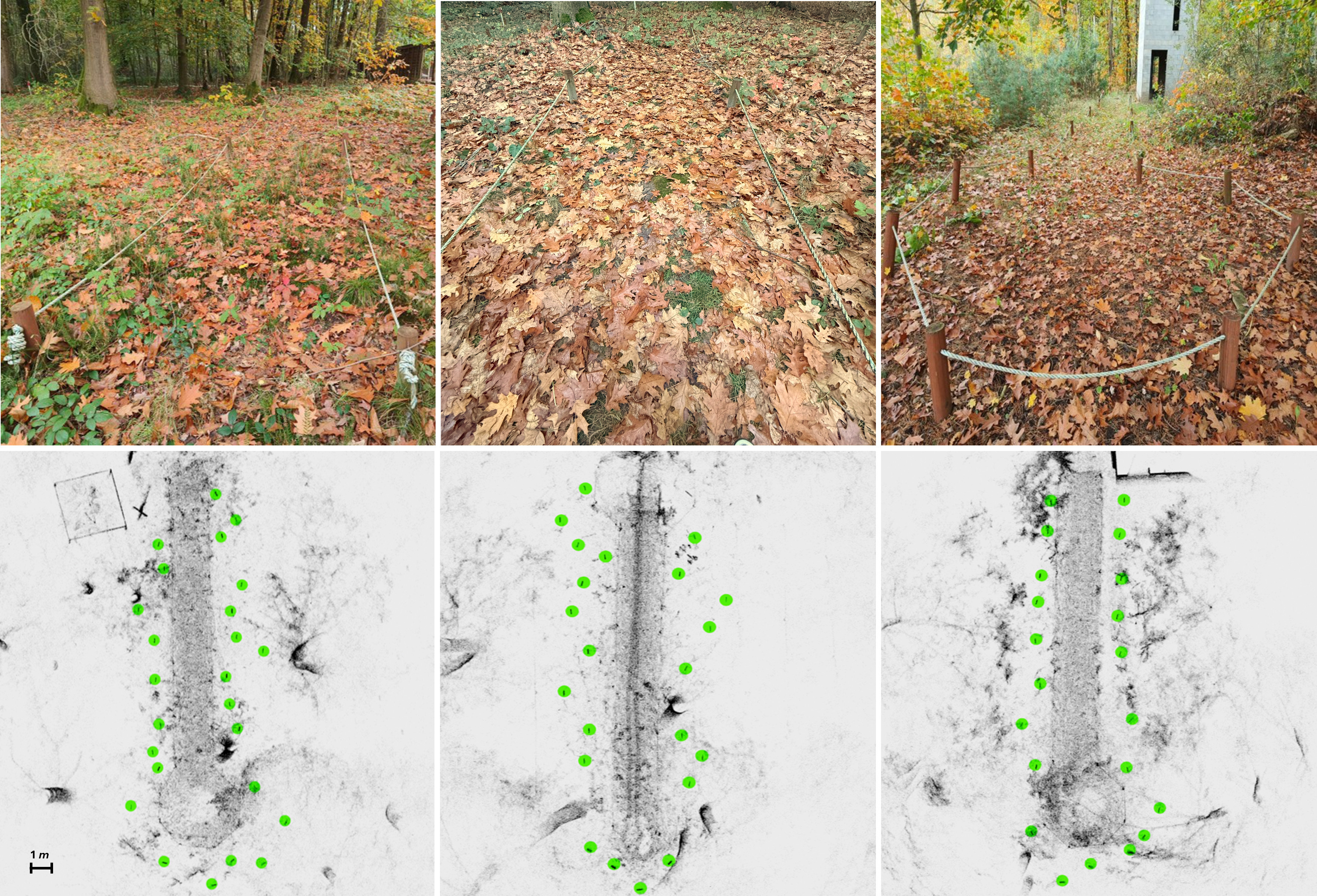}%
        };

        % -- Use a scope to map coordinates from 0..1 in x and y
        \begin{scope}[x={(image.south east)}, y={(image.north west)}]

            % ============== Top Row Labels (Horizontal) ==============
            \node[anchor=south] at (0.17,1.01) {\textbf{Track 1}};
            \node[anchor=south] at (0.50,1.01) {\textbf{Track 2}};
            \node[anchor=south] at (0.83,1.01) {\textbf{Track 3}};

            % ============== Vertical Labels for Each Row ==============
            % For the top row (roughly near the center of the top row)
            \node[rotate=90, anchor=center] at (-0.03,0.75) {\textbf{Vegetation Overview}};

            % For the bottom row (roughly near the center of the bottom row)
            \node[rotate=90, anchor=center] at (-0.03,0.25) {\textbf{Top-View Point Cloud}};

            % You can fine-tune these coordinates to match your exact layout
        \end{scope}
    \end{tikzpicture}
    \caption{Overview of the three tracks in the dataset. The first row displays an image of each track to show the condition of the vegetation. The second row presents a top-view point cloud representation, highlighting the locations of targets and illustrating their distribution across the tracks. Scale: \SI{1}{m} (bottom left).}
    \label{fig:all_tracks}
\end{figure*}
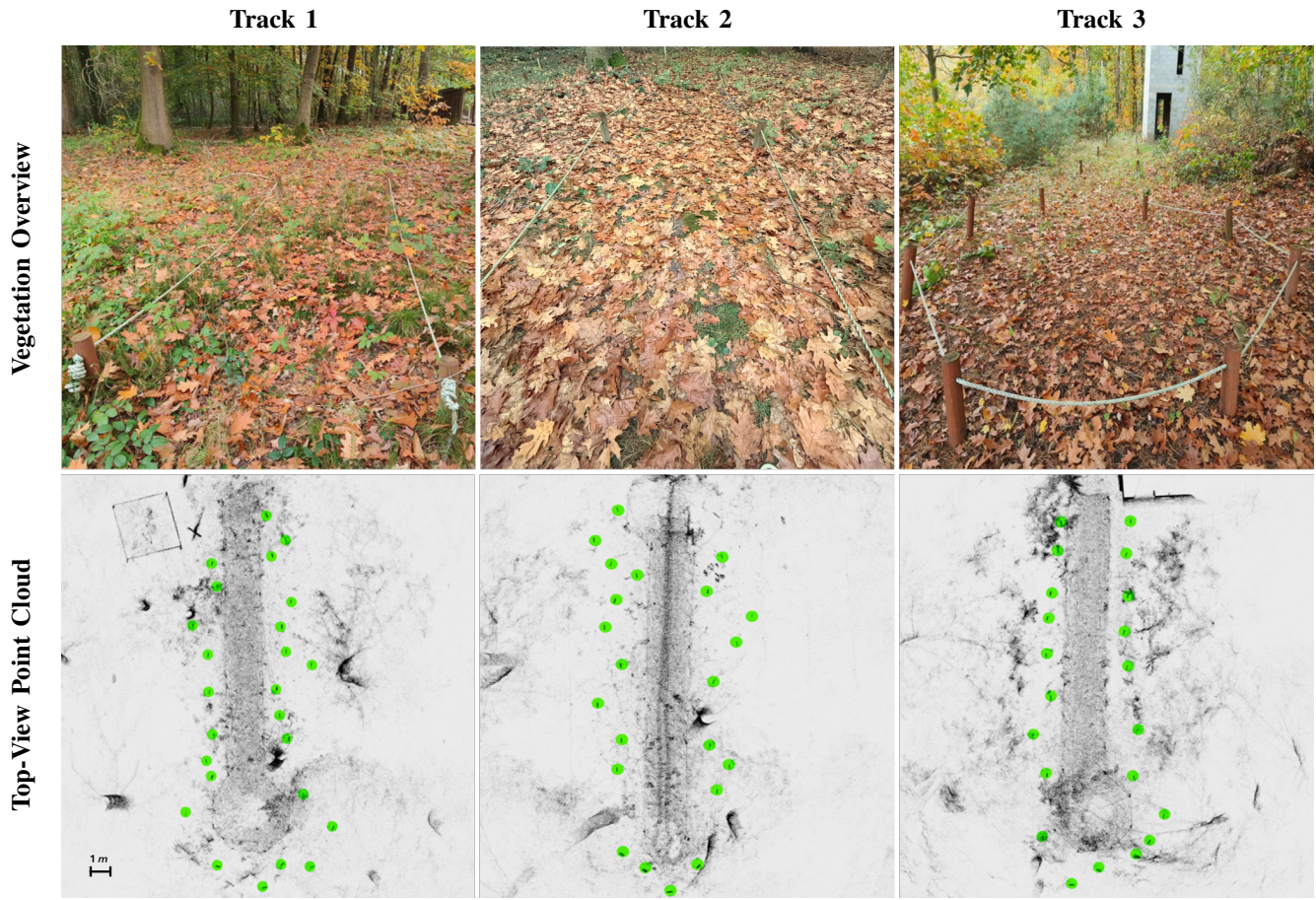
%%%%%%%%%%%%%%%%%%%%%%%%%
In this section, we present the MineInsight dataset~\cite{mineinsight_git}. We describe the targets, detail the environmental conditions with varying vegetation and lighting, and outline the recorded data sequences, including the UGV and robotic arm configurations. Finally, we explain the reference data generation process, which combines an automated pipeline with human supervision to improve the quality of the annotations.
\label{sec_dataset}
%%%%%%%%%%%%%%%%%%%%%%%%%
\renewcommand{\vec}[1]{\boldsymbol{#1}}
\newcommand{\mat}[1]{\mathbf{#1}}
\newcommand{\matB}{\mat{B}}
\newcommand{\matC}{\mat{C}}
\newcommand{\matW}{\mat{W}}
\newcommand{\matV}{\mat{V}}
\newcommand{\SE}{\textnormal{SE}}
%%%%%%%%%%%%%%%%%%%%%%%%%
\subsection{Targets}
\label{subsec_targets}
We placed a combination of AP and AT inert landmines and common items that might be found in natural environments across three tracks to construct a diverse dataset.
The items were left overnight, allowing their temperatures to equilibrate with the ambient environment and better mimic realistic conditions.
All the items included in the dataset are shown in Figure~\ref{fig:all_targets}.
We summarize the distribution of landmines and common items per track as follows:
%%%%%%%%%%%%%%%%%%%%%%%%%

\noindent\textbf{Track~1} (25 targets): 15 landmines (11 AP, 4 AT) and 10 common items.
\noindent\textbf{Track~2} (21 targets): 11 landmines (6 AP, 5 AT) and 10 common items.  
\noindent\textbf{Track~3} (21 targets): 11 AP landmines and 10 common items.  
%%%%%%%%%%%%%%%%%%%%%%%%%
\subsection{Environment}
\label{subsec_environment}
We selected three outdoor environments for data sampling, each consisting of an approximately \SI{15}{m} long and \SI{2.5}{m} wide track designated for inspection. The track boundaries were marked by ropes to indicate restricted access. The terrain was primarily covered with leaves, and the campaign took place under consistently cloudy conditions (no direct sunlight). We categorize the tracks as follows:
\begin{itemize}
    \item \textbf{Track 1}: Forest floor with low to mid-high vegetation, mostly covered by freshly fallen deciduous leaves over moist clay-loam soil. Recorded between 13:00 and 14:15, with ambient air temperatures ranging from \SI{12.8}{\degree C} to \SI{13.6}{\degree C}. Conditions were overcast, giving diffuse light and a uniform thermal background;
    \item \textbf{Track 2}: Low to mid-high vegetation, with a more heterogeneous leaf litter than Track~1. The surface shows uneven layers of leaves with occasional gaps exposing the moist clay-loam soil beneath. Recorded between 15:15 and 16:05, with ambient air temperatures between \SI{12.5}{\degree C} and \SI{13.2}{\degree C}. Moisture variations created visible contrasts across modalities, with thermal differences between leaf-covered and exposed ground;
    \item \textbf{Track 3}: Denser and taller vegetation compared to the previous tracks, with soil partly visible under uneven leaf cover. Recorded between 17:30 and 18:00, under twilight conditions, with ambient air temperatures between \SI{12.8}{\degree C} and \SI{13.4}{\degree C}. Cooling soil and foliage created mixed thermal patterns. The RGB camera is unavailable for this track due to a hardware failure, although the twilight conditions would have limited its usefulness.
\end{itemize}
An overview of the vegetation conditions for each track, as well as the distribution of landmines and common items, is provided in Figure~\ref{fig:all_tracks}. In addition, we release a minute-by-minute climatology log covering both the target placement phase and the campaign itself. The log records atmospheric variables (temperature, humidity, wind, pressure, cloud coverage, precipitation) as well as soil temperatures at multiple depths, providing the environmental context needed to interpret the thermal data.
\subsection{Sequences}
\label{subsec_sequences}
For each lane, we recorded three sequences: the reference sequence and two evaluation sequences.
We use the reference sequence only for reference data generation, as detailed in section~\ref{subsec_ref_data_gen}. 
\noindent The first and second evaluation sequences, referred to as Sequence 1 and Sequence 2, differ in the position of the robotic arm during data sampling.
%%%%%%% FIGURE %%%%%%%
\begin{figure}[t]
\centering
    \includegraphics[width=1\linewidth]{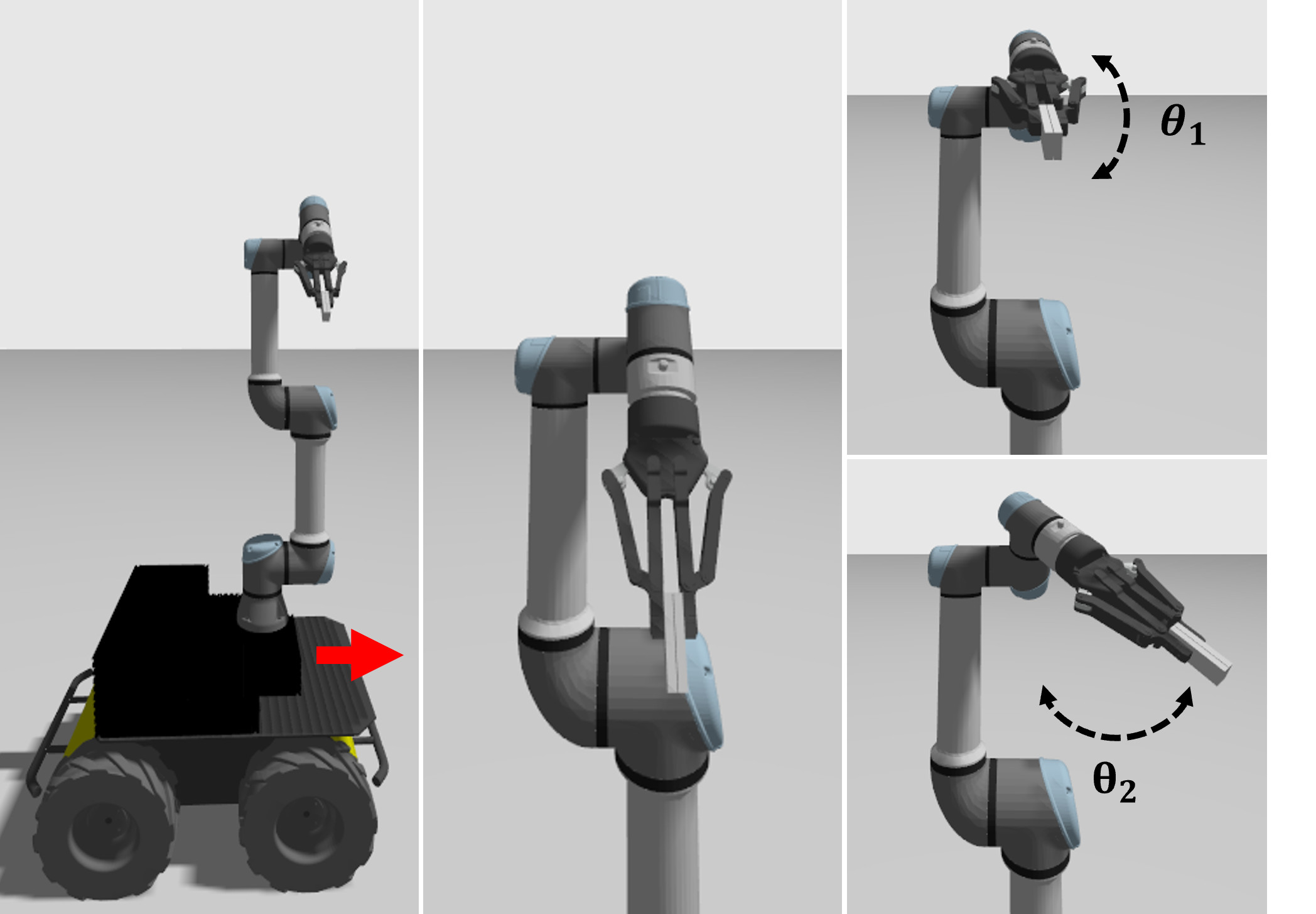}
    \caption{\textit{Left -} Arm stow and UGV motion: The manipulator is positioned relative to the UGV’s body, with the vehicle’s forward motion direction indicated by the red arrow.
\textit{Center -} End effector position in the first sequence: The arm is extended to show the initial placement and orientation of the end effector.
\textit{Right -} End effector position in the second sequence: The rotations of the arm’s wrist joints are highlighted. The angles $\theta_1$ and $\theta_2$ correspond to the rotational movements of wrist joint 1 and wrist joint 2, respectively.}
\label{fig:ugv_arm_position}
\end{figure}
%%%%%%%%%%%%%%%%%%%%%%%%
Figure~\ref{fig:ugv_arm_position} illustrates the arm's position relative to the UGV, along with the rotation of wrist joint 1 and wrist joint 2 for each sequence.
\noindent{In \textbf{Sequence 1}, the UGV traverses the entire track back and forth while maintaining its position on the right side, at an average speed of \SI{0.2}{m/s}. The robotic arm is fully extended vertically, with $\theta_1$ tilted 45° downward. The sequence lasts about 3 to 4 minutes, depending on the track length.}
\noindent{In \textbf{Sequence 2}, the UGV traverses the entire track back and forth while maintaining its position on the right side, at an average speed of \SI{0.2}{m/s}. At approximately every meter, it stops to scan the environment on both sides using the robotic arm. Including these stops, the average speed for scanning the entire track is \SI{0.054}{m/s}.}
The robotic arm remains fully extended in a vertical position, with both wrist joints 1 and 2 actuated. The end effector sweeps from left to right, with wrist joint 1 moving between $\theta_1$~=~34° and $\theta_1$~=~70°, and wrist joint 2 moving between $\theta_2$~=~55° and $\theta_2$~=~130°. The sequence lasts between 19 and 22 minutes, depending on the track.
\noindent In total, we have recorded six sequences, providing a total recording duration of approximately one hour. The reported frame counts correspond only to the sensors in the robotic arm sensor suite and are approximate: around 38,000 frames from the RGB camera, 53,000 frames from the VIS-SWIR camera, and 108,000 frames from the LWIR camera.
%%%
\begin{figure}[t]
   \centering
    \includegraphics[width=1\linewidth]{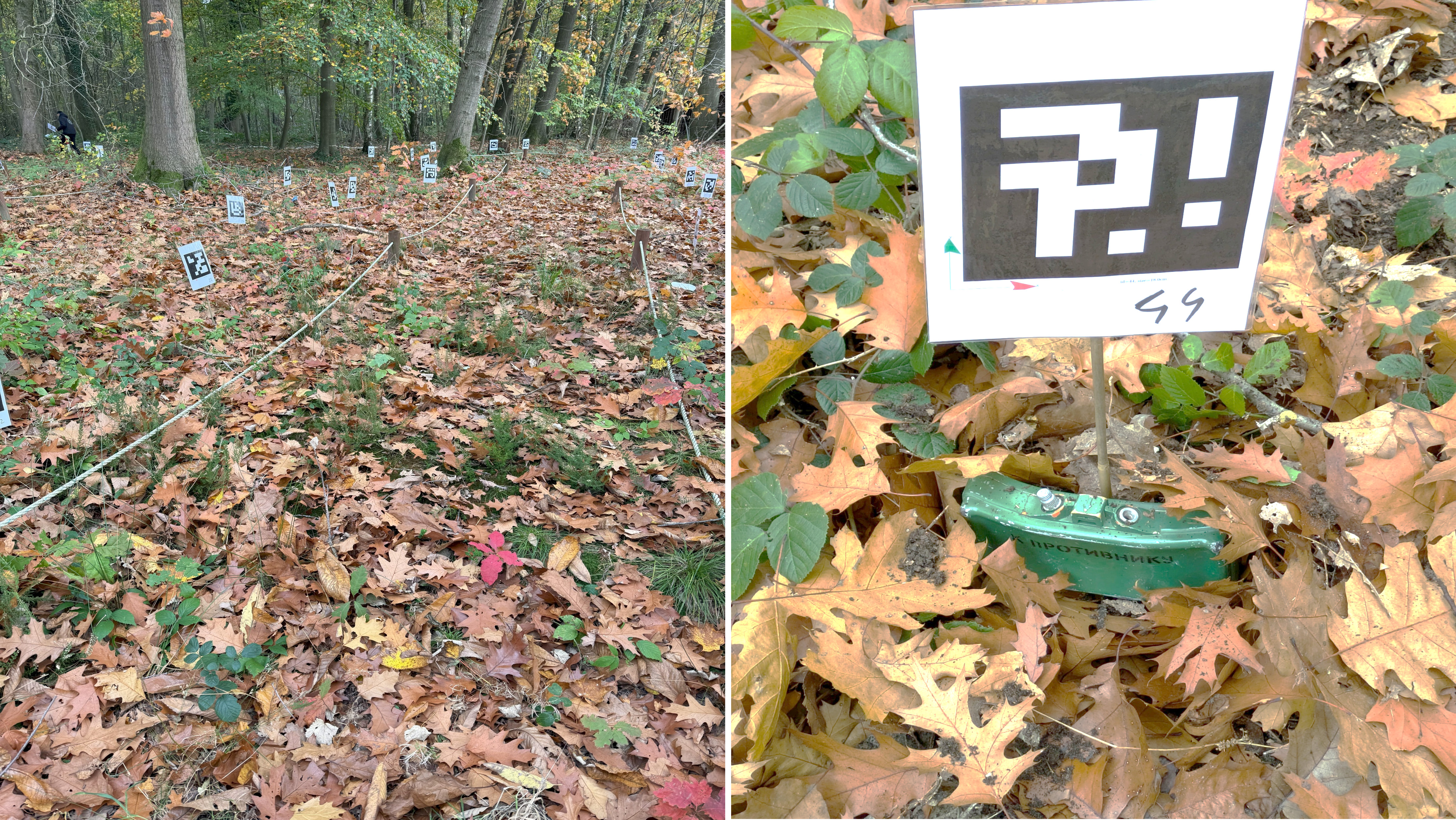}
    \caption{\textit{Left -} The track is ready to sample the reference sequence. \textit{Right -} A MON-50 landmine and its corresponding AprilTag. The pose of each AprilTag, denoted as \(A_i\), is evaluated, allowing determination of the position of the corresponding target \(T_i\).}
    \label{fig:april_tags}
\end{figure}
%%%%% FIGURE BEGINS %%%%%
\begin{figure*}[t]
    \centering
    \begin{tikzpicture}
        \node[anchor=south west, inner sep=0] (image) at (0,0)
            {\includegraphics[width=0.75\linewidth]{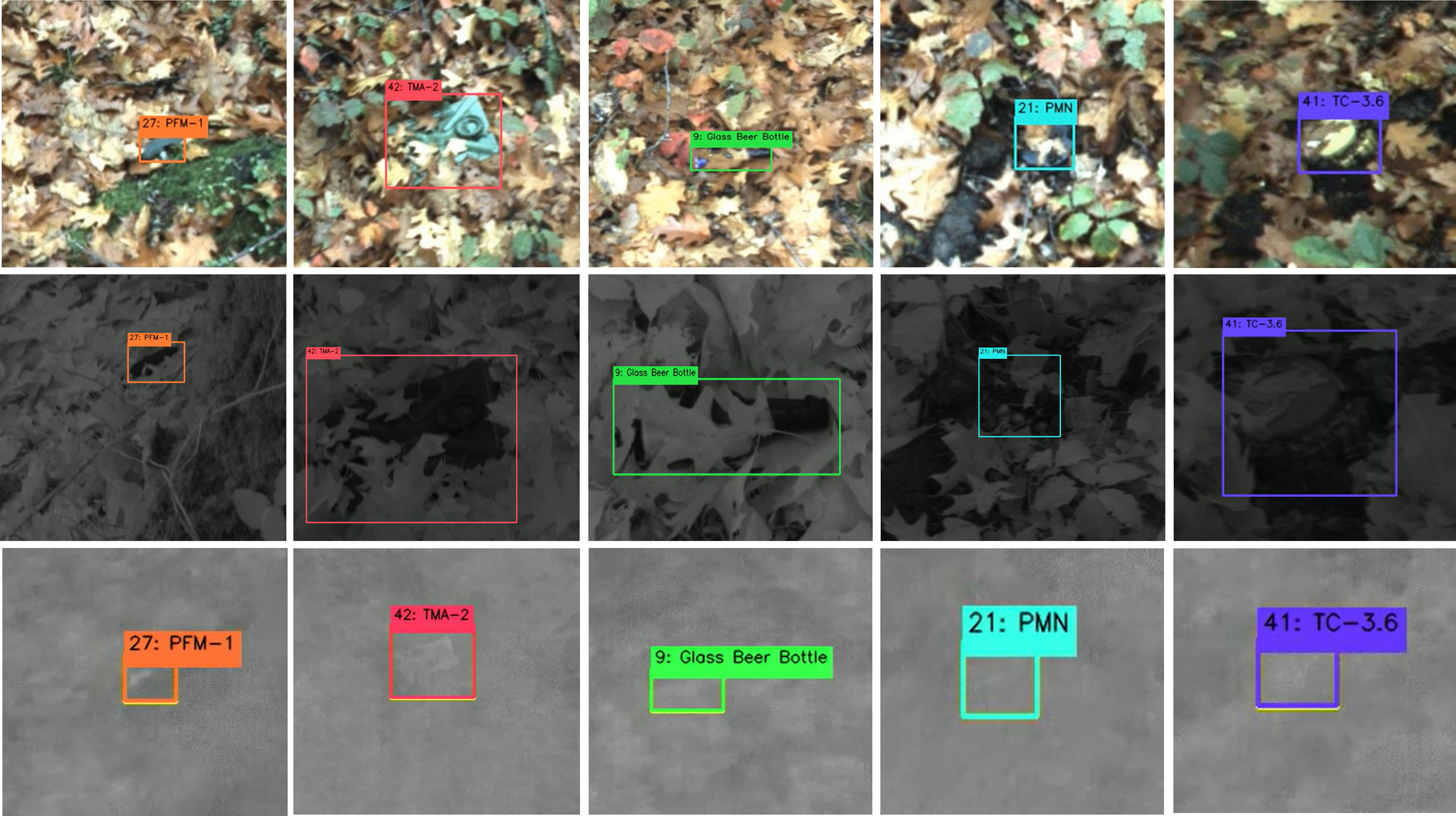}};
        
        \begin{scope}[x={(image.south east)}, y={(image.north west)}]
            \node[anchor=south] at (0.10,1.02) {\textbf{PFM-1}};
            \node[anchor=south] at (0.30,1.02) {\textbf{TMA-2}};
            \node[anchor=south] at (0.50,1.02) {\textbf{Glass beer bottle}};
            \node[anchor=south] at (0.70,1.02) {\textbf{PMN}};
            \node[anchor=south] at (0.90,1.02) {\textbf{TC-3.6}};
            
            \node[anchor=south, rotate=90] at (-0.02,0.82) {\textbf{RGB}};
            \node[anchor=south, rotate=90] at (-0.02,0.50) {\textbf{VIS-SWIR}};
            \node[anchor=south, rotate=90] at (-0.02,0.18) {\textbf{LWIR}};
        \end{scope}
    \end{tikzpicture}
    \caption{Examples of five targets from the dataset, each captured by the robotic arm’s sensor suite in RGB, VIS–SWIR, and LWIR. Bounding boxes are obtained from the annotation pipeline and refined through human supervision to better match the visible extent of each object.}

    \label{fig:all_the_pictures_of_the_dataset}
\end{figure*}
%%%%% FIGURE ENDS %%%%%
%%%%%%%%%%%%%%%%%%%%%%%
\subsection{Reference data generation}
\label{subsec_ref_data_gen}
As illustrated in Figure~\ref{fig:all_the_pictures_of_the_dataset}, targets can hardly be detected by the human eye in the images. Direct manual labeling is difficult, as many targets are barely visible in the raw imagery. Therefore, we implement an annotation pipeline that generates initial estimates using AprilTags~\cite{apriltag}, SLAM~\cite{nguyen2023slict}, ICP, and sensor kinematics, which are then adjusted by human intervention.
We placed an AprilTag at each target location on every track and recorded a reference sequence. An example of a track prepared for the reference sequence, along with a sample target and its corresponding tag, is shown in Figure~\ref{fig:april_tags}. 
The main steps of the reference data generation process are the following:
%%%
\begin{enumerate}
%%% ITEM 1 %%%
\item {\textbf{Target positioning in the reference sequence coordinate frame:}} We detect AprilTags across all frames using the Sevensense Core Research module. With accurate intrinsic calibration of the camera (\( C_{\text{SEV}} \)), we determine each detected AprilTag's pose, including both position and orientation, relative to the camera frame. We denote this pose as \( {}^{C_{\text{SEV}}}\mat{T}_{A_i} \in \SE(3) \), where \( i \) indexes the detected AprilTag and \( \SE(3) \) represents the Special Euclidean group, which describes rigid body transformations in 3D space. Here, \( C_{\text{SEV}} \) refers to the Sevensense camera's coordinate frame.

Using the extrinsic calibration between the camera, the platform's LiDAR sensor (Livox Mid-360), and the associated IMU, we transform the AprilTag pose from the camera frame into the IMU coordinate frame as follows:
\begin{align}
    {}^I\mat{T}_{T_i} = {}^I\mat{T}_{L} \, {}^L\mat{T}_{C_{\text{SEV}}} \, {}^{C_{\text{SEV}}}\mat{T}_{A_i}\,
    {}^{A_i}\mat{T}_{T_i}
\end{align}

where \( {}^I\mat{T}_{L} \) and \( {}^L\mat{T}_{C_{\text{SEV}}} \) denote the rigid body transformations from the LiDAR to the IMU and from the camera to the LiDAR, respectively.
Additionally, \({}^{A_i}\mat{T}_{T_i}\) represents the transformation from the top left corner of the AprilTag to the actual target on the ground.
The known offset along the stick plus a ground-plane intersection estimate is used to determine the actual target location from the AprilTag’s pose.

We associate each image where a target is detected with the corresponding pose from the SLAM system, defined as the pose of the IMU in the world coordinate frame of the reference sequence, denoted \( {}^{W_R}\mat{T}_{I} \). Consequently, we express the target poses in the reference sequence coordinate frame as:
\begin{align}
    {}^{W_R}\mat{T}_{T_i} = {}^{W_R}\mat{T}_{I} \, {}^I\mat{T}_{T_i}
\end{align}
%%% ITEM 2 %%%
\item \textbf{Computing the transformation between the reference sequence coordinate frame and the evaluation sequence coordinate frame:} 

We execute a point-to-plane ICP algorithm on the point clouds generated by running the SLAM system on both the reference sequence and the evaluation sequence. This process estimates the transformation from the reference sequence coordinate system (\( W_R \)), to the evaluation sequence coordinate frame \( W_E \). 

We denote this transformation as \( {}^{W_E} \mat{T}_{W_R} \).

%%% ITEM 3 %%%
\item \textbf{Transforming the target locations into the evaluation sequence coordinate frame:}

After computing the transformation \( {}^{W_E} \mat{T}_{W_R} \), we apply it to the target locations initially expressed in the reference coordinate frame \( W_R \). This step allows us to represent the target positions within the evaluation sequence coordinate frame \( W_E \). 

The transformation of each target location is computed as:

\begin{equation}
    {}^{W_E} \mat{T}_{T_i} = {}^{W_E} \mat{T}_{W_R} \, {}^{W_R} \mat{T}_{T_i}
\end{equation}

Here, \( {}^{W_E} \mat{T}_{T_i} \) represents the pose of the \( i \)-th target in the evaluation sequence coordinate frame, while \( {}^{W_R} \mat{T}_{T_i} \) denotes its corresponding pose in the reference frame.
%%% ITEM 4 %%%
\item \textbf{Transforming into the UGV and camera coordinate frames:}

To express the target poses in the UGV coordinate frame along its trajectory and in the camera coordinate frame, we apply the inverse transformations from Step 1, now within the evaluation sequence.

First, we transform the target pose from the evaluation sequence frame \( W_E \) to the IMU coordinate frame \( I \):

\begin{align}
    {}^{I} \mat{T}_{T_i} = \left( {}^{W_E} \mat{T}_{I} \right)^{-1} \, {}^{W_E} \mat{T}_{T_i}
\end{align}

In the case of the RGB camera, we reproject the target into the camera coordinate frame \( C_{\text{RGB}} \) as:

\begin{align}
    {}^{C_{\text{RGB}}} \mat{T}_{T_i} = \left( {}^I \mat{T}_{L} \, {}^L \mat{T}_{C_{\text{RGB}}} \right)^{-1} \, {}^I \mat{T}_{T_i}
\end{align}

where \( {}^I \mat{T}_{L} \) represents the transformation from LiDAR to IMU, and \( {}^L \mat{T}_{C_{\text{RGB}}} \) corresponds to the transformation from the RGB camera to the LiDAR frame. It incorporates the conversion from the RGB camera frame to the robot base and then to the Livox Mid-360 frame, including the necessary arm kinematics.

Finally, using the intrinsic and extrinsic calibration parameters of the respective camera, we project the transformed target locations onto the image plane to obtain the pixel coordinates of the center of each target, and apply a fixed-size bounding box according to the corresponding camera resolution.
\item \textbf{Human-supervised adjustment of bounding boxes:} 
The automatically generated boxes serve as an initial guess.
Annotators adjust them to better match the visible extent of each target, mitigating errors propagated from AprilTag detection, pose estimation, or perspective distortions. 
RGB and VIS–SWIR annotations are manually revised. For LWIR, where signatures are often faint, adjustments rely on reprojections from the corrected RGB and VIS–SWIR annotations, with human input.

\end{enumerate}

%% file: 5_baseline_evaluation.tex
\section{Baseline Evaluation}
\label{sec:baseline_evaluation}

As a baseline, we trained YOLOv8~\cite{yolov8} on the SULAND dataset~\cite{vivoli_yolo}, 
which is the closest publicly available dataset to our scenario. 
Even with extensive augmentation strategies, the model did not transfer successfully to MineInsight and failed to produce reliable detections. 
Rather than serving as a performance benchmark, this outcome highlights the domain gap between existing datasets and MineInsight, driven by its clutter-rich environments with vegetation and distractor objects, as well as a broader variety of landmine types.
This underlines MineInsight’s value as a benchmark for studying domain adaptation and multi-modal fusion in realistic demining scenarios.

%% file: 6_limitations.tex
\section{Limitations}
\label{sec:limitations}

MineInsight is, to the best of our knowledge, the first multi-modal dataset for landmine detection with RGB, VIS–SWIR, and LWIR sensing. Its current release, however, has some constraints. All data were collected during a single late-autumn campaign under overcast conditions. As a result, models trained on this dataset may face domain gaps when applied to different seasons, terrains, or weather conditions, such as sunny, snowy, arid, or grassy environments. The release also lacks RGB data for Track~3, though twilight conditions would in any case have limited its usefulness. While MineInsight surpasses existing public datasets in scale and sensor diversity, the number of targets and total recording time remain modest relative to real-world scenarios. More broadly, the dataset highlights the domain-adaptation challenges faced when transferring models trained elsewhere to MineInsight. Finally, annotations are adjusted with human supervision; minor inaccuracies may remain, particularly for faint thermal signatures. These limitations motivate the future extensions outlined in the conclusion.

%% file: 7_conclusion.tex
\section{Conclusion}
We introduced MineInsight, a multi-sensor, multi-spectral dataset for robotic landmine detection in off-road environments. 
It combines dual viewpoints (UGV and robotic arm), multi-spectral imaging (monochrome, RGB, VIS–SWIR, LWIR), dual LiDARs, and synchronized data to address occlusions and sensor limitations. 
The dataset includes bounding boxes generated by an automated pipeline with a human-in-the-loop process to support the evaluation of detection algorithms, and it can also serve generic off-road robotics tasks such as navigation and mapping. In addition, the dataset includes minute-by-minute climatology measurements, which provide valuable context for exploiting LWIR data and studying thermal contrast under varying conditions.
As noted in Section~\ref{sec:limitations}, the current release is limited to a single late-autumn campaign and may introduce domain gaps under other environmental conditions. 
Future work will expand MineInsight with recordings across varied terrains, climates, and seasons, additional sensors, new tracks, and further landmine types. 
We also aim to explore spectral fusion, landmine segmentation, and synthetic data for domain adaptation. 
Finally, the issue of RGB degradation at night, observed in Track~3, could be further studied with lighting systems to extend visibility.

\addtolength{\textheight}{-2.5cm}   % This command serves to balance the column lengths

%%%%%%%%%%%%%%%%%%%%%%%%%%%%%%%%%%%%%%%%%%%%%%%%%%%%%%%%%%%%%%%%%%%%%%%%%%%%%%%%

%%%%%%%%%%%%%%%%%%%%%%%%%%%%%%%%%%%%%%%%%%%%%%%%%%%%%%%%%%%%%%%%%%%%%%%%%%%%%%%%

%%%%%%%%%%%%%%%%%%%%%%%%%%%%%%%%%%%%%%%%%%%%%%%%%%%%%%%%%%%%%%%%%%%%%%%%%%%%%%%%

%% file: paper_bib.bib
@article{nevliudov_robots,
    author = {Nevliudov, Igor and Yanushkevych, Dmytro and Ivanov, Leonid},
    year = {2021},
    pages = {47--52},
    title = {Analysis of the state of creation of robotic complexes for humanitarian demining},
    volume = {6},
    journal = {Technology audit and production reserves}
}

@inproceedings{popov_uav,
    author={Popov, Mykhailo O. and Stankevich, Sergey A. and Mosov, Sergey P. and Titarenko, Olga V. and Topolnytskyi, Maksym V. and Dugin, Stanislav S.},
    booktitle={IEEE International Conference on Smart Technologies (EUROCON)}, 
    title={Landmine Detection with {UAV}-based Optical Data Fusion}, 
    year={2021},
    pages={175--178}
}

@article{colreavy_cnn,
    author = {Colreavy-Donnelly, Simon and Caraffini, Fabio and Kuhn, Stefan and Gongora, Mario and Florez, Johana and Parra, Carlos},
    year = {2020},
    title = {Shallow Buried Improvised Explosive Device Detection Via Convolutional Neural Networks},
    volume = {27},
    number = {4},
    pages = {403--416},
    journal = {Integrated Computer-Aided Engineering}
}

@inproceedings{winfred_thermal_cnn,
    author={Lin, Winfred and Wang, Qiang},
    booktitle={IEEE International Conference on Imaging Systems and Techniques (IST)}, 
    title={{MineSeeker}: A Novel Design for Detecting Explosive Mines Using Low-Resolution Thermal Imaging}, 
    year={2024},
    pages={1--5},
}

@article{milan_uav_uxo_det,
    author = {Bajić, Milan and Potočnik, Božidar},
    title = {{UAV} Thermal Imaging for Unexploded Ordnance Detection by Using Deep Learning},
    journal = {Remote Sensing},
    volume = {15},
    year = {2023},
    number = {4: 967},
}

@article{baur_occlusion,
    author = {Baur, Jasper and Dewey, Kyle and Steinberg, Gabriel and Nitsche, Frank O.},
    title = {Modeling the Effect of Vegetation Coverage on Unmanned Aerial Vehicles-Based Object Detection: A Study in the Minefield Environment},
    journal = {Remote Sensing},
    volume = {16},
    year = {2024},
    number = {12: 2046},
}

@inproceedings{rellis3d_dataset,
    author={Jiang, Peng and Osteen, Philip and Wigness, Maggie and Saripalli, Srikanth},
    booktitle={IEEE International Conference on Robotics and Automation (ICRA)}, 
    title={{RELLIS-3D} Dataset: Data, Benchmarks and Analysis}, 
    year={2021},
    pages={1110--1116}
}

@inproceedings{rugd_dataset,
    author={Wigness, Maggie and Eum, Sungmin and Rogers, John G. and Han, David and Kwon, Heesung},
    booktitle={IEEE/RSJ International Conference on Intelligent Robots and Systems (IROS)}, 
    title={A {RUGD} Dataset for Autonomous Navigation and Visual Perception in Unstructured Outdoor Environments}, 
    year={2019},
    pages={5000--5007},
}

@inproceedings{citrus_dataset,
    title={Multimodal Dataset for Localization, Mapping and Crop Monitoring in Citrus Tree Farms},
    author={Teng, Hanzhe and Wang, Yipeng and Song, Xiaoao and Karydis, Konstantinos},
    booktitle={International Symposium on Visual Computing (ISVC)},
    pages={571--582},
    year={2023}
}

@inproceedings{extrinsic_calib,
    author={Koide, Kenji and Oishi, Shuji and Yokozuka, Masashi and Banno, Atsuhiko},
    booktitle={IEEE International Conference on Robotics and Automation (ICRA)}, 
    title={General, Single-shot, Target-less, and Automatic {LiDAR}-Camera Extrinsic Calibration Toolbox}, 
    year={2023},
    pages={11301--11307},
 }

@inproceedings{kalibr_1,
  author={Rehder, Joern and Nikolic, Janosch and Schneider, Thomas and Hinzmann, Timo and Siegwart, Roland},
  booktitle={IEEE International Conference on Robotics and Automation (ICRA)}, 
  title={Extending kalibr: Calibrating the extrinsics of multiple {IMU}s and of individual axes}, 
  year={2016},
  pages={4304--4311}
}

@inproceedings{kalibr_2,
    author={Furgale, Paul and Rehder, Joern and Siegwart, Roland},
    booktitle={IEEE/RSJ International Conference on Intelligent Robots and Systems (IROS)}, 
    title={Unified temporal and spatial calibration for multi-sensor systems}, 
    year={2013},
    pages={1280--1286}
}

@article{thermal_calib,
    title = {Highly accurate geometric calibration for infrared cameras using inexpensive calibration targets},
    author = {R. Usamentiaga and D.F. Garcia and C. Ibarra-Castanedo and X. Maldague},
    journal = {Measurement},
    volume = {112},
    pages = {105--116},
    year = {2017}
}

@inproceedings{dnn_landmine,
    author = {Mohamed Fikry, Refaat and Kasban, Hany},
    title = {Deep Neural Networks for Landmines Images Classification},
    booktitle={International Conference on Advanced Intelligent Systems and Informatics (AISI)},
    pages = {126--136},
    year = {2020}
}

@article{vivoli_yolo,
    author = {Vivoli, Emanuele and Bertini, Marco and Capineri, L.},
    year = {2024},
    title = {Deep Learning-Based Real-Time Detection of Surface Landmines Using Optical Imaging},
    volume = {16},
    number = {4: 677},
    journal = {Remote Sensing},
}

@article{baur_dl_uav,
    author = {Baur, Jasper and Steinberg, Gabriel and Nikulin, Alex and Chiu, Kenneth and de Smet, Timothy S.},
    title = {Applying Deep Learning to Automate {UAV}-Based Detection of Scatterable Landmines},
    journal = {Remote Sensing},
    volume = {12},
    year = {2020},
    number = {5: 859},
}

@electronic{scatterable_1_7,
    author       = {de Smet, Timothy and Nikulin, Alex and Baur, Jasper},
    title        = {Scatterable Landmine Detection Project Datasets 1-7},
    year         = {2020},
    institution  = {Geological Sciences and Environmental Studies Faculty Scholarship},
    howpublished = {\url{https://orb.binghamton.edu/geology_fac/4}},
    note         = {Accessed: \today}
}

@electronic{scatterable_8_9,
    author       = {Steinberg, Gabriel and Baur, Jasper and Nikulin, Alex and de Smet, Timothy},
    title        = {Scatterable Landmine Detection Project Datasets 1-7},
    year         = {2020},
    institution  = {Geological Sciences and Environmental Studies Faculty Scholarship},
    howpublished = {\url{https://orb.binghamton.edu/geology_fac/11}},
    note         = {Accessed: \today}
}

@electronic{scatterable_10_22,
    author       = {Baur, Jasper and De Smet, Tim and Nikulin, Alex and Steinberg, Gabriel},
    title        = {Scatterable Landmine Detection Project Datasets 10-24},
    year         = {2020},
    institution  = {Geological Sciences and Environmental Studies Faculty Scholarship},
    howpublished = {\url{https://orb.binghamton.edu/geology_fac/15}},
    note         = {Accessed: \today}
}

@ARTICLE{kaya_thermal,
    author={Kaya, Serkan and Leloglu, Ugur Murat},
    journal={IEEE Journal of Selected Topics in Applied Earth Observations and Remote Sensing}, 
    title={Buried and Surface Mine Detection From Thermal Image Time Series}, 
    volume={10},
    number={10},
    year={2017},
    pages={4544--4552}}

@article{leloglu_dataset,
    doi = {10.21227/tg8m-6f29},
    author = {Leloglu, Ugur Murat and Kaya, Serkan},
    journal = {IEEE Dataport},
    title = {Landmine Thermal Image Series},
    year = {2019} 
}

@article{tamayo_dataset,
    author = {Tenorio Tamayo, Alejandro and Forero Ramirez, Juan and Garcia, Bryan and Loaiza-Correa, Humberto and Restrepo-Girón, Andrés and Nope Rodríguez, Sandra and Barandica-López, Asfur and Buitrago, Jose Tomas},
    year = {2023},
    volume = {49},    
    pages = {109443},
    title = {Dataset of thermographic images for the detection of buried landmines},
    journal = {Data in Brief},
}

@electronic{roboflow_web,
    author= "Dwyer, B. and Nelson, J. and Hansen, T. et al.",
    title = "{Roboflow (Version 1.0)}",
    howpublished={Available: \url{https://roboflow.com/}},
    note = "Accessed: \today",
    year  = 2024,
}

@inproceedings{apriltag,
    author={Olson, Edwin},
    booktitle={IEEE International Conference on Robotics and Automation (ICRA)}, 
    title={{AprilTag}: A robust and flexible visual fiducial system}, 
    year={2011},
    pages={3400--3407},
}

@article{nguyen2023slict,
    title         = {{SLICT}: Multi-input Multi-scale Surfel-Based {LiDAR}-Inertial Continuous-Time Odometry and Mapping},
    author        = {Nguyen, Thien-Minh and Duberg, Daniel and Jensfelt, Patric and Yuan, Shenghai and Xie, Lihua},
    journal       = {IEEE Robotics and Automation Letters},
    volume        = {8},
    number        = {4},
    pages         = {2102--2109},
    year          = {2023},
    publisher     = {IEEE}
}

@article{Schramm2021,
    author  = {Schramm, Sebastian and Rangel, Johannes and Aguirre Salazar, Daniela and Schmoll, Robert and Kroll, Andreas},
    title   = {Multispectral Geometric Calibration of Cameras in Visual and Infrared Spectral Range},
    journal = {IEEE Sensors},
    year    = {2021},
    volume  = {21},
    number  = {2},
    pages   = {2159--2168},
}

@electronic{mineinsight_git,
    author= "Malizia, Mario and Hamesse, Charles and Hasselmann, Ken and De Cubber, Geert and Tsiogkas, Nikolaos and Demeester, Eric and Haelterman, Rob",
    title = "{{MineInsight}: A Multi-sensor Dataset for Humanitarian Demining
Robotics in Off-Road Environments - GitHub Repository}",
    howpublished={Available: \url{https://github.com/mariomlz99/MineInsight}},
    note = "Accessed: \today",
    year  = 2025,
}

@misc{yolov8,
  author = {Jocher, G. and Chaurasia, A. and Qiu, J.},
  title = {{YOLO} by Ultralytics},
  year = {2023},
  note = {Available: \url{https://github.com/ultralytics/ultralytics}}
}
